\documentclass[10pt,twocolumn,letterpaper]{article}

\usepackage[pagenumbers]{iccv} %
\usepackage{bm}
\usepackage{mathtools}
\usepackage{multirow}

\usepackage{colortbl}

\definecolor{firstcolor}{rgb}{1, 0.6, 0.6}
\definecolor{secondcolor}{rgb}{1, 0.8, 0.6}
\definecolor{thirdcolor}{rgb}{1,1, 0.6}

\newcommand{\fst}[1]{\cellcolor{firstcolor}#1}
\newcommand{\snd}[1]{\cellcolor{secondcolor}#1}
\newcommand{\trd}[1]{\cellcolor{thirdcolor}#1}

\definecolor{iccvblue}{rgb}{0.21,0.49,0.74}
\usepackage[pagebackref,breaklinks,colorlinks,allcolors=iccvblue]{hyperref}

\def\confName{ICCV}
\def\confYear{2025}

\title{NormalCrafter: Learning Temporally Consistent Normals \\ from Video Diffusion Priors}

\author{
Yanrui Bin$^{1}$
\quad 
Wenbo Hu$^{2\star}$ 
\quad 
Haoyuan Wang$^{3}$
\quad 
Xinya Chen$^{4}$ 
\quad
Bing Wang$^{1\dag}$ \\
	$^{1}$Spatial Intelligence Group, The Hong Kong Polytechnic University \quad $^{2}$ARC Lab, Tencent PCG \\ $^{3}$City University of Hong Kong $^4$Huazhong University of Science and Technology \\
    {\tt\small \url{https://normalcrafter.github.io/}}\\
}

\begin{document}
\twocolumn[{%
		\renewcommand\twocolumn[1][]{#1}%
		\maketitle
            \vspace{-2em}
		\begin{center}

             \setlength{\tabcolsep}{0pt}
             \def\mywidth{.33}
             \begin{tabular}{ccc}
             \includegraphics[width=\mywidth\linewidth]{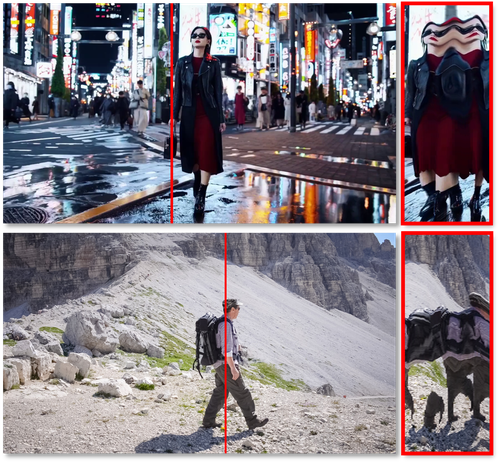} \hspace{0.1mm} &
              \includegraphics[width=\mywidth\linewidth]{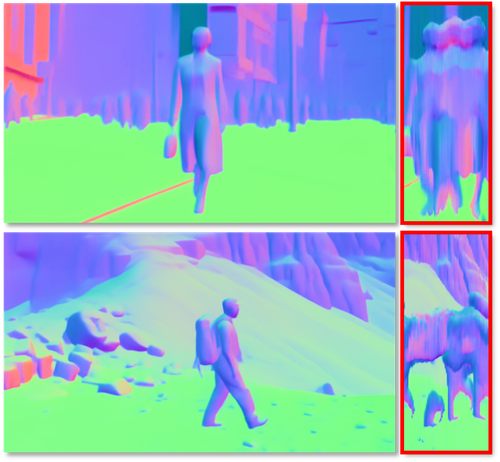} \hspace{0.1mm}&
              \includegraphics[width=\mywidth\linewidth]{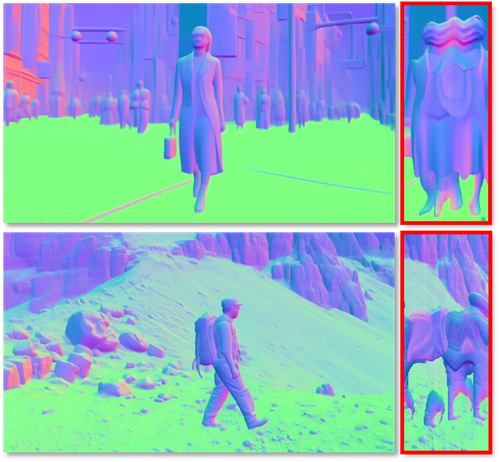}\\
            
              \begin{small}Input\end{small} &
              \begin{small}Marigold-E2E-FT\end{small} &
              \begin{small}Ours\end{small} 
              
             \end{tabular}
     
\vspace{-0.2cm}			
   \captionsetup{type=figure}
			\captionof{figure}{
					We innovate NormalCrafter, a novel video normal estimation model, that can generate temporally consistent normal sequences with fine-grained details from open-world videos with arbitrary lengths. 
					Compared to results from state-of-the-art image normal estimators, Marigold-E2E-FT~\cite{E2E}, our results exhibit both higher spatial fidelity and temporal consistency, as shown in the frame visualizations and temporal profiles (marked by the red lines and rectangles).
					}
			\label{fig:teaser}
		\end{center}
	}
    ]

\maketitle
\footnotetext[0]{$^{\star}$ Project leader. 
\quad $^\dag$~Corresponding author.
}

\begin{abstract}
Surface normal estimation serves as a cornerstone for a spectrum of computer vision applications.
While numerous efforts have been devoted to static image scenarios, ensuring temporal coherence in video-based normal estimation remains a formidable challenge.
Instead of merely augmenting existing methods with temporal components, we present NormalCrafter to leverage the inherent temporal priors of video diffusion models.
To secure high-fidelity normal estimation across sequences, we propose Semantic Feature Regularization (SFR), which aligns diffusion features with semantic cues, encouraging the model to concentrate on the intrinsic semantics of the scene.
Moreover, we introduce a two-stage training protocol that leverages both latent and pixel space learning to preserve spatial accuracy while maintaining long temporal context.
Extensive evaluations demonstrate the efficacy of our method, showcasing a superior performance in generating temporally consistent normal sequences with intricate details from diverse videos.
\end{abstract}
    
\section{Introduction}
\label{sec:intro}

\begin{figure*}[t]
     \setlength{\tabcolsep}{0pt}
     \def\mywidth{.2}
     \begin{tabular}{ccccc}
     \includegraphics[width=\mywidth\linewidth]{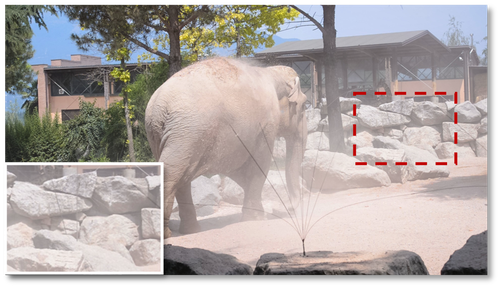} &
      \includegraphics[width=\mywidth\linewidth]{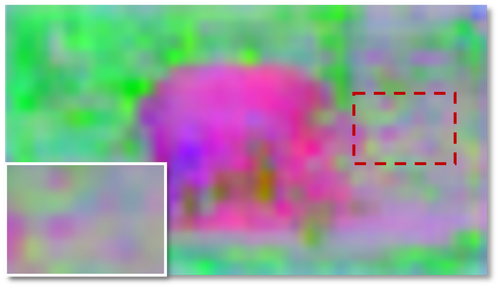}
      &
      \includegraphics[width=\mywidth\linewidth]{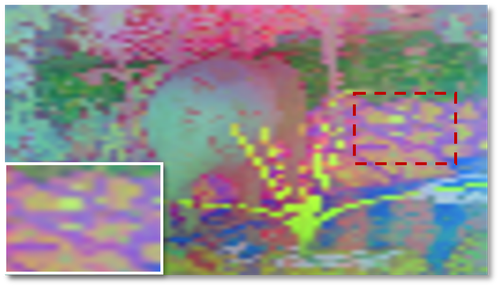}
      &
      \includegraphics[width=\mywidth\linewidth]{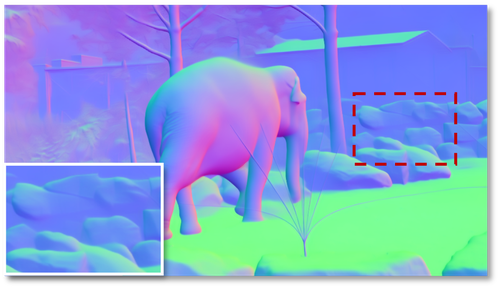} &
      \includegraphics[width=\mywidth\linewidth]{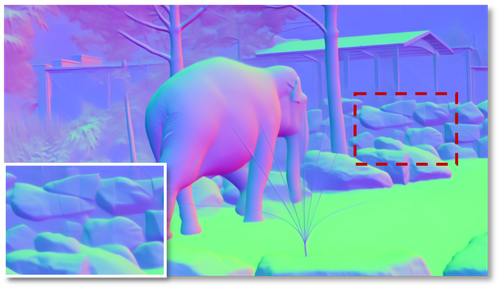}\\
    
      \begin{small}RGB\end{small} &
      \begin{small}SVD Feature\end{small} &
      \begin{small}DINO Feature\end{small} &
      \begin{small}Ours w/o SFG \end{small} &
      \begin{small}Ours\end{small} 
      
     \end{tabular}\vspace{-0.1cm}
     \caption{
            Naively repurposing video diffusion models, \eg SVD~\cite{blattmann2023stable}, for normal estimation (Ours w/o SFG) produces over-smoothed predictions, due to insufficient high-level semantic cues in SVD features. 
            By leveraging Semantic Feature Regularization (SFR) to align diffusion features with DINO~\cite{caron2021emerging}, our approach yields sharper and more fine-grained normal predictions.
        }
   \label{fig:motivation}
\end{figure*}

Surface normals, as pivotal descriptors of 3D scene geometry, underpin a spectrum of applications, including 3D reconstruction, relighting, video editing, and mixed reality.
Estimating high-fidelity and temporally consistent normals from diverse, unconstrained videos remains a formidable challenge, owing to variations in scene layouts, illuminations, camera motions, and scene dynamics.

Recent advancements in normal estimation from monocular images have embraced both discriminative~\cite{bae2024rethinking, omnidata_v2, eftekhar2021omnidata, bae2021estimating, do2020surface, wang2020vplnet} and generative paradigms~\cite{geowizard, stablenormal, lotus, E2E, GenPercept}. 
While discriminative approaches remain hampered by the limitations of training data scale and quality, resulting in suboptimal zero-shot generalization,
generative methods harness pre-trained diffusion priors to deliver state-of-the-art performance on open-world images, even when confined to synthetic training data.
However, these methods are inherently designed for static imagery, neglecting the temporal dynamics of videos and consequently inducing temporal inconsistency or flickering, as demonstrated in \cref{fig:teaser}.

In this paper, we propose NormalCrafter, a novel video normal estimation model that generates temporally consistent normal sequences exhibiting rich, fine-grained details from unconstrained open-world videos of arbitrary lengths.
Rather than incrementally incorporating temporal layers or devising complex stabilization schemes for image-based normal estimators, we harness the potential of video diffusion models for a more robust approach to video normal estimation.
Although repurposed video diffusion models have achieved remarkable success in depth estimation, normal estimation presents its own set of challenges, particularly in preserving the high-frequency, semantics-driven details inherent in surface normals.
Naively applying video diffusion models to normal estimation often yields suboptimal performance, such as the over-smoothing of normal predictions, as illustrated in~\cref{fig:motivation}.
To this end, we introduce a \emph{semantic feature regularization} (SFR) technique that directs the model's focus toward the semantics by aligning diffusion features with semantic representations extracted from an external encoder, \eg, DINO~\cite{caron2021emerging}.
Furthermore, recent findings demonstrate that supervising the final output of the variational autoencoder (VAE) in image-based depth or normal estimation, rather than operating solely in the latent space, significantly enhances spatial fidelity.
However, this direct supervision considerably increases GPU memory consumption during training, as it requires expanding the compact latent space into the high-dimensional pixel space, thereby restricting training to shorter video clips.
To address this issue, we propose a two-stage training strategy: first, training the full model in the latent space to effectively capture long-term temporal context, and then fine-tuning the spatial layers in the pixel space to improve spatial accuracy while preserving the capacity for long sequence inference.

We perform a comprehensive evaluation of our NormalCrafter across a wide range of datasets under zero-shot settings. 
Both qualitative and quantitative analyses reveal that NormalCrafter attains state-of-the-art performance in open-world video normal estimation, significantly surpassing existing methodologies. 
Moreover, our rigorous ablation experiments substantiate the effectiveness of the proposed semantic feature regularization and two-stage training strategy in enhancing both the spatial fidelity and temporal consistency of normal predictions.
Our contributions are summarized below:
\begin{itemize}
    \item  We introduce NormalCrafter, a novel framework that generates temporally consistent normal sequences with intricate, fine-grained details for open-world videos of arbitrary lengths, outperforming existing approaches by a substantial margin.
    \item  We propose the semantic feature regularization (SFR) technique, which directs the model's focus towards meaningful semantics by aligning diffusion features with high-level semantic representations.
    \item  We devise a two-stage training strategy that leverages both latent and pixel-space supervision, enabling the generation of normal sequences with long temporal context while preserving high spatial accuracy.
\end{itemize}

\section{Related Work}
Our method relates to two primary research streams: video diffusion models and surface normal estimation. The latter can be categorized into discriminative methods that directly regress normal maps from the input, and more recent diffusion-based approaches that leverage the priors of generative diffusion models for this task.

\begin{figure*}[!t]
  \centering
  \includegraphics[width=1.\linewidth]{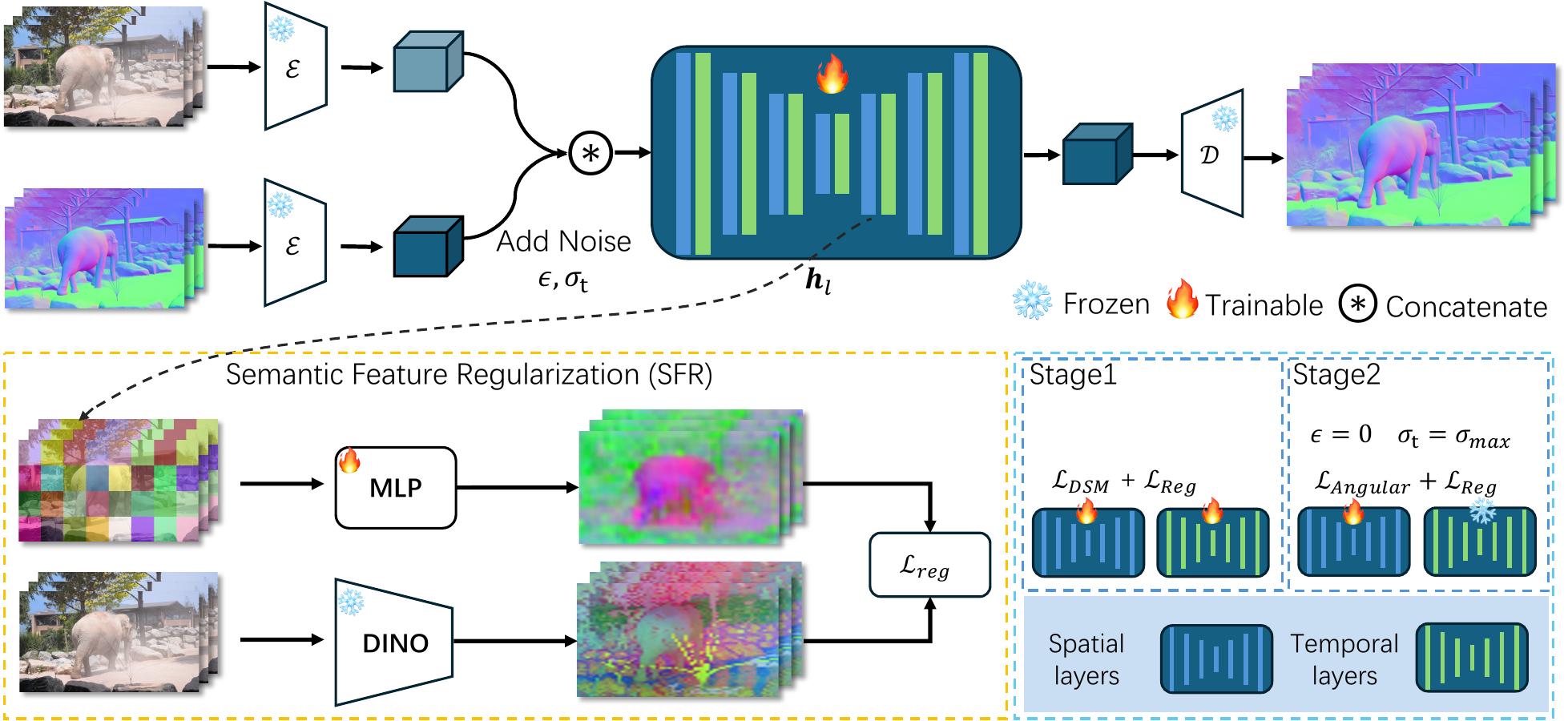}\vspace{-0.0cm}
  \caption{
      \textbf{Overview of our NormalCrafter}. 
      We model the video normal estimation task with a video diffusion model conditioned on input RGB frames. 
      We propose Semantic Feature Regularization (SFR) $\mathcal{L}_{\text{reg}}$ to align the diffusion features with robust semantic representations from DINO encoder, encouraging the model to concentrate on the intrinsic semantics for accurate and detailed normal estimation.
      Our training protocol consists of two stages: 1) training the entire U-Net in the latent space with diffusion score matching $\mathcal{L}_{\text{DSM}}$ and SFR $\mathcal{L}_{\text{reg}}$; 2) fine-tuning only the spatial layers in pixel space with angular loss $\mathcal{L}_{\text{angular}}$ and SFR $\mathcal{L}_{\text{reg}}$.
      }
    \label{fig:pipeline}
\vspace{-0.3cm}
\end{figure*}

\noindent\textbf{Discriminative surface normal estimation.}
Surface normal estimation has been studied for decades. Early work~\cite{hoiem2005automatic, fouhey2013data, fouhey2014unfolding} used hand-crafted features with learning-based classification, exemplified by~\cite{hoiem2005automatic} that discretized normals.
With deep learning, convolutional neural networks (CNNs) drastically improved this task. Wang \etal~\cite{wang2020vplnet} combined CNNs with vanishing point analysis. Do \etal~\cite{do2020surface} introduced a spatial rectifier to align tilted images with high-likelihood training distributions. Bae \etal~\cite{bae2021estimating} leveraged aleatoric uncertainty for improved robustness and performance in small structures. Eftekhar \etal~\cite{eftekhar2021omnidata} compiled over 12 million images from diverse scenes and camera intrinsics, training a U-Net on this massive dataset. Its successor Omnidata v2~\cite{omnidata_v2} utilized a transformer-based model with advanced 3D augmentation and cross-task consistency. Recently, DSINE~\cite{bae2024rethinking} achieved state-of-the-art results by incorporating per-pixel ray directions and modeling relationships between neighboring normals, providing a strong baseline for our approach.

\noindent\textbf{Diffusion-based surface normal estimation.}
Recently, pre-trained diffusion models have gained strong attention. Marigold~\cite{marigold} fine-tuned Stable Diffusion (SD)~\cite{SD} for dense prediction tasks conditioned on images.
Concurrently, Geowizard~\cite{geowizard} fine-tuned SD to output both depth and normal maps. Although effective, these models required iterative denoising, causing high computational overhead.
To address this, some works~\cite{GenPercept, E2E} replaced multi-step denoising with a single-step approach, sacrificing detailed geometry. Addressing this trade-off, Lotus~\cite{lotus} added an image reconstruction objective to enhance details, while StableNormal~\cite{stablenormal} used a coarse-to-fine scheme for sharper results.
Despite their strong priors, these methods overlook temporal context and often produce flickering artifacts in videos. 
Concurrent with our work, BufferAnytime~\cite{BufferAnytime} augmented Marigold-E2E-FT~\cite{E2E} with temporal layers, using optical-flow-based supervision to stabilize results.
However, optical flow alone cannot guarantee correct normal correspondences in consecutive frames, as it overlooks camera motion and scene dynamics.
In contrast, our approach learns video normal estimation directly from large-scale labeled data and pre-trained diffusion priors, delivering a comprehensive spatio-temporal understanding of the scene.
As they neither release the model nor the evaluation data, we exclude it from comparisons.

\noindent\textbf{Video diffusion model.}
Recent advances in video generation increasingly rely on diffusion models~\cite{ho2020denoising, sohl2015deep, song2020score} to synthesize temporally coherent frames conditioned on text or images. 
Latent Diffusion Models (LDMs)~\cite{SD} offer improved efficiency by operating in a compressed latent space, enabling high-resolution image generation with reduced computational cost. 
Building on LDMs, Blattmann \etal~\cite{blattmann2023align} added temporal convolution and attention layers to SD, training these on video data. 
Stable Video Diffusion (SVD)~\cite{blattmann2023stable} further refined this approach with extensive training strategy and curated video data. 
SVD produces high-quality videos and serves as a model prior for diverse video-related tasks~\cite{hu2024depthcrafter, shao2024learning}. 
In this paper, we leverage the rich spatio-temporal priors of SVD for high-fidelity, consistent video normal estimation.

\section{Method}

We present NormalCrafter, a reliable video normal estimator derived from video diffusion models (VDMs).%
The overall pipeline of NormalCrafter is illustrated in~\cref{fig:pipeline}.
Given a video $\bm{c} \in \mathbb{R}^{F \times W \times H \times 3}$ with frame number $F$, our objective is to generate normal estimations $\bm{n} \in \mathbb{R}^{F \times W \times H \times 3}$ that are spatially accurate and temporally consistent.

\subsection{Normal Estimator with VDMs}

To alleviate computational overhead, modern VDMs typically operate in a compressed latent space by leveraging a Variational Autoencoder (VAE) for efficient encoding and decoding of video frames.
Since normal maps share the same dimensions as RGB image frames, we seamlessly utilize the same VAE for both the normal maps $\bm{n}$ and the corresponding video $\bm{c}$:
\begin{gather}
    \label{eq:vae}
    \bm{z}^{x} = \mathcal{E}(\bm{x}), \quad \hat{\bm{x}} = \mathcal{D}(\bm{z}^x),
\end{gather}
where $\mathcal{E}$ and $\mathcal{D}$ denote the encoder and decoder of the VAE, respectively, $\bm{x}$ may represent either $\bm{n}$ or $\bm{c}$, and $\hat{\bm{x}}$ is the reconstructed counterpart of $\bm{x}$.
However, most existing VAEs are pre-trained on RGB frames, which is suboptimal for normal maps.
Therefore, we specifically fine-tune the VAE decoder on normal data to bolster the reconstruction quality.

\vspace{0.5em}
\noindent\textbf{Diffusion-based normal estimation.}
In diffusion framework, normal estimation is formulated as a transformation between a simple noise distribution to a target data distribution $p(\bm{z}^n \, | \, \bm{z}^c)$ conditioned on the input video latents $\bm{z}^c$.
On the one hand, to map $p(\bm{z}^n \, | \, \bm{z}^c)$ into the noise distribution, a forward diffusion sequence is applied by injecting Gaussian noise with variance $\sigma^2_t$ into the latent normal sequence $\bm{z}^n_0$ at each time step $t$:
\begin{gather}
    \bm{z}^n_t = \bm{z}^n_0 + \sigma^2_t \bm{\epsilon}, \quad \bm{\epsilon} \sim \mathcal{N}(\bm{0}, \bm{I}),
\end{gather}
where $\bm{z}^n_t \sim p(\bm{z}^n; \sigma_t)$ denotes the noisy latent normal sequence.
When $\sigma_t$ becomes sufficiently large, the noisy latent distribution $p(\bm{z}^n; \sigma_t)$ becomes statistically indistinguishable from a pure Gaussian prior.
On the other hand, to transform the noise distribution to $p(\bm{z}^n \, | \, \bm{z}^c)$, a reverse denoising process begins by drawing a noise sample $\bm{\epsilon} \sim \mathcal{N}(\bm{0}, \sigma^2_{\text{max}}\bm{I})$ and iteratively transforms it into $\bm{z}^n_0$ through a learned denoiser $D_{\theta}$.
This denoiser is trained via denoising score matching (DSM)~\cite{blattmann2023stable}:
\begin{gather}
    \label{eq:latent_loss}
    \mathcal{L}_{\text{DSM}} \coloneqq
    \mathbb{E}_{\bm{z}^n \sim p(\bm{z}^n; \sigma_t), \sigma_t \sim p(\sigma)} \lambda(\sigma_t) \lVert D_{\theta}(\bm{z}^n_t; \sigma_t; \bm{z}^c) - \bm{z}^n_0 \rVert^2_2,
\end{gather}
where $p(\sigma)$ is the noise level distribution during training, and $\lambda(\sigma_t) = (1+\sigma^2_t)\sigma^{-2}_t$ is the weight function.
The denoiser function $D_{\theta}$ specifies the noise-level distribution during training.
We build our NormalCrafter on top of the SVD model, which is originally designed for generating videos from an input image.
We adapt this SVD framework into our NormalCrafter model by substituting the image input with a frame-wise concatenation of the noisy normal latent $\bm{z}^n_t$ and the conditional video latent $\bm{z}^c$, as shown in~\cref{fig:pipeline}.

\subsection{Semantic Feature Regularization}

SVD was originally designed for conditioning on a single input image, and may therefore struggle to effectively accumulate contextual information when extended to sequences of multiple frames.
As illustrated in~\cref{fig:motivation}, the initial SVD intermediate features exhibit semantic ambiguity; for instance, the stone region in the background is over-blurred, contradicting the detailed geometry evident in the original frames.
As a result, leveraging SVD directly leads to over-smooth normal maps, lacking the intricate structural details in the corresponding areas.
To verify whether high-level semantic representations can preserve such geometric details, we visualized the features of the DINO encoder~\cite{caron2021emerging} by applying PCA~\cite{PCA}.
As shown in ~\cref{fig:motivation}, the DINO features exhibit a strong correlation with the geometric structures of the input frames, exemplified by their refined representations of both stone and plant regions.

This motivates us to incorporate semantic features into the diffusion model to further elevate the quality of normal estimation.
To this end, the most straightforward approach is to augment the diffusion model with DINO features as an additional conditioning.
However, such a design leads to substantial computational and memory overheads during training and inference.
Therefore, we propose a Semantic Feature Regularization (SFR) method to rectify the semantic ambiguities in SVD features by aligning them with robust semantic representations throughout training, inspired by REPA~\cite{yu2024representation}.
This alignment encourages the diffusion model to concentrate on the intrinsic semantics of the input frames, yielding more accurate and finely detailed normal maps.
Moreover, SFR introduces overhead solely during training, leaving inference unaltered with no extra costs.

Specifically, as shown in~\cref{fig:pipeline}, we initially derive the DINO features $\bm{h}_{\text{dino}} = f(\bm{c}) \in \mathbb{R}^{N \times D}$ from the input video frames $\bm{c}$, where $N$ and $D$ indicate the number of patches and the embedding dimension, respectively.
Then, we extract the intermediate features $\bm{h}_{l}$ from the $l$-th layer of the diffusion model, and project them into the DINO feature space using a learnable multilayer perceptron $h_{\phi}$.
Finally, we regularize the projected features to align with the DINO features by maximizing the patch-wise cosine similarities:
\begin{gather}
    \label{eq:sfr}
    \mathcal{L}_{\text{reg}}(\theta, \phi) \coloneqq -\mathbb{E}_{\bm{c}} \Big[ \frac{1}{N}\sum_{n=1}^{N} \mathrm{cossim}(\bm{h}_{\text{dino}}^{[n]}, h_{\phi}(\bm{h}_{l}^{[n]}) \Big],
\end{gather}
where $n$ is the patch index, and $\mathrm{cossim}$ is the cosine similarity function between two vectors.

\subsection{Two-Stage Training Protocol}

Although training NormalCrafter in the latent space with the loss $\mathcal{L}_{\text{DSM}} + \mathcal{L}_{\text{reg}}$ is feasible, it may not yield optimal results in terms of accuracy or efficiency, as highlighted in~\cite{E2E}.
Instead, it proposes to fine-tune the image diffusion model in a single end-to-end step for depth and normal estimation, directly optimizing the pixel-wise loss in the image space, thereby achieving superior spatial fidelity alongside improved efficiency.
However, extending such an approach to video normal estimation heavily restricts the length of training clips, since it requires employing VAE to decode the latent normal sequence into pixel space to compute the loss, which drastically elevates memory requirements, especially for long sequences.

To this end, we propose a two-stage training protocol that artfully balances the need for long temporal context modeling with high-precision spatial fidelity.
As shown in~\cref{fig:pipeline}, we first train NormalCrafter in the latent space under the combined objectives of $\mathcal{L}_{\text{DSM}} + \mathcal{L}_{\text{reg}}$.
The sequence length in this stage is randomly sampled from $[1, 14]$, enabling NormalCrafter to flexibly adapt to diverse video durations.
Moreover, this setup facilitates training on both single-frame and multi-frame video datasets.
In the second stage, we fine-tune only the spatial layers by decoding the latent normal sequence into pixel space and employing the loss $\mathcal{L}_{\text{angular}} + \mathcal{L}_{\text{reg}}$.
Here, $\mathcal{L}_{\text{angular}}$ is defined as:
\begin{equation}
\label{eq:angular_loss}
    \mathcal{L}_{\text{angular}} = \frac{1}{HW} \sum_{i,j} \arccos \left( \frac{n^*_{i,j} \cdot \hat{n}_{i,j}}{\|n^*_{i,j}\| \|\hat{n}_{i,j}\|} \right),
\end{equation}
where $n^*_{i,j}$ is the ground-truth normal at pixel $(i,j)$, and $\hat{n}_{i,j}$ is the predicted normal.
During this second stage, the sequence length is randomly sampled from $[1, 4]$ frames, thereby easing GPU memory constraints.
Since the model has already absorbed long-range temporal cues in the first stage, and only the spatial layers are refined in the second, this two-stage protocol allows the model to enjoy the benefits of end-to-end fine-tuning while preserving its capacity to process extensive sequences.

\section{Experiment}
\begin{table*}[t]
\caption{
\textbf{Quantitative evaluations.}
The top section shows the results on single-image benchmarks, while the bottom section shows the results on video benchmarks. ``mean'' and ``med'' denote the mean and median angular error, respectively. 
The last column shows the average ranking across all metrics. 
The \colorbox{firstcolor}{best}, \colorbox{secondcolor}{2nd-best}, and \colorbox{thirdcolor}{3rd-best} results are highlighted.
}
\label{table:benchmark}
\centering
\footnotesize
\setlength\tabcolsep{4pt}
\resizebox{\linewidth}{!}{
\begin{tabular}{l|cc|ccc|c|cc|ccc|c}
\toprule
\multirow{2}{*}{Method} 
& \multicolumn{6}{c|}{NYUv2~\cite{nyu} (Single-image Benchmark) }
& \multicolumn{6}{c}{iBims~\cite{ibim} (Single-image Benchmark)} \\
\cline{2-13}
& mean $\downarrow$ & med $\downarrow$ & {\scriptsize $11.25^{\circ} \uparrow$} & {\scriptsize $22.5^{\circ} \uparrow$} & {\scriptsize $30^{\circ} \uparrow$} & Rank $\downarrow$
& mean $\downarrow$ & med $\downarrow$ & {\scriptsize $11.25^{\circ} \uparrow$} & {\scriptsize $22.5^{\circ} \uparrow$} & {\scriptsize $30^{\circ} \uparrow$} & Rank $\downarrow$ \\
\midrule
DSINE~\cite{bae2024rethinking} & 
16.4 & 8.4 & 59.6 & 77.7 & 83.5 & 4.8 &
17.1 & \trd{6.1} & 67.4 & 79.0 & 82.3 & 5.0\\
GeoWizard~\cite{geowizard} & 
18.6 & 12.0 & 46.4 & 76.1 & 83.0 & 7.0 &
20.5 & 10.9 & 51.5 & 75.2 & 80.1 & 7.0 \\
GenPercept~\cite{GenPercept} & 
16.4 & \trd{8.0} & \snd{60.9} & \snd{78.3} & 83.7 & \trd{3.2} &
\trd{16.3} & 6.3 & \snd{69.5} & \bf\fst{81.1} & \snd{84.1} & \snd{2.4} \\
StableNormal~\cite{stablenormal} & 
17.7 & 10.3 & 54.2 & \trd{78.1} & \snd{84.1} & 4.6 &
17.0 & 7.0 & 68.0 & \snd{80.9} & \bf\fst{84.2} & 3.4\\
Lotus-D~\cite{lotus} & 
\snd{16.2} & 8.4 & 59.8 & 78.0 & \trd{83.9} & 3.4 &
17.1 & 6.8 & 66.4 & 79.4 & 83.0 & 5.2\\
Marigold-E2E-FT~\cite{E2E} & 
\snd{16.2} & \bf\fst{7.6} & \bf\fst{61.4} & 77.9 & 83.5 & \snd{2.8} &
\bf\fst{15.8} & \bf\fst{5.5} & \bf\fst{69.9} & \trd{80.6} & \trd{83.9} & \bf\fst{1.8} \\
\hline
Ours & 
\bf\fst{15.4} & \snd{7.9} & \bf\fst{61.4} & \bf\fst{79.4} & \bf\fst{85.1} & \bf\fst{1.2} & 
\snd{16.1} & \snd{5.9} & \trd{68.9} & 80.4 & 83.7 & \trd{3.0} \\
\midrule
\multirow{2}{*}{\small Method} 
& \multicolumn{6}{c|}{Scannet~\cite{scannet} (Video Benchmark)}
& \multicolumn{6}{c}{Sintel~\cite{sintel} (Video Benchmark)} \\
\cline{2-13}
& mean $\downarrow$ & med $\downarrow$ & {\scriptsize $11.25^{\circ} \uparrow$} & {\scriptsize $22.5^{\circ} \uparrow$} & {\scriptsize $30^{\circ} \uparrow$} & Rank $\downarrow$
& mean $\downarrow$ & med $\downarrow$ & {\scriptsize $11.25^{\circ} \uparrow$} & {\scriptsize $22.5^{\circ} \uparrow$} & {\scriptsize $30^{\circ} \uparrow$} & Rank $\downarrow$ \\
\midrule
DSINE~\cite{bae2024rethinking} & 
15.5 & 8.0 & 62.4 & 79.5 & 84.9 & 5.4 & 
34.9 & 28.1 & \trd{21.5} & 41.5 & 52.7 & 4.6 \\
GeoWizard~\cite{geowizard} & 
18.9 & 13.1 & 41.7 & 75.1 & 83.0 & 7.0 & 
37.6 & 32.0 & 11.7 & 32.8 & 46.8 & 6.4 \\
GenPercept~\cite{GenPercept} & 
14.5 & 7.2 & \trd{66.0} & \trd{81.8} & \trd{86.7} & \trd{3.4} & 
34.6 & \trd{26.2} & 18.4 & \trd{43.8} & \trd{55.8} & \trd{3.6} \\
StableNormal~\cite{stablenormal} & 
15.9 & 10.0 & 57.0 & \snd{81.9} & \snd{87.0} & 4.4 &
38.8 & 32.7 & 17.9 & 36.1 & 46.6 & 6.6 \\
Lotus-D~\cite{lotus} & 
\trd{14.3} & \trd{7.1} & 65.6 & 81.4 & 86.5  & 3.8 & 
\snd{32.3} & \snd{25.5} & \snd{22.4} & \snd{44.9} & \snd{57.0} & \snd{2.0} \\
Marigold-E2E-FT~\cite{E2E} & 
\snd{14.1} & \bf\fst{6.3} & \bf\fst{67.6} & 81.7 & 86.4 & \snd{2.6} & 
\trd{33.5} & 27.0 & \trd{21.5} & 43.0 & 54.3 & \trd{3.6} \\
\hline
Ours & 
\bf\fst{13.3} & \snd{6.8} & \snd{67.4} & \bf\fst{82.9} & \bf\fst{87.9} & \bf\fst{1.4} &
\bf\fst{30.7} & \bf\fst{23.9} & \bf\fst{23.5} & \bf\fst{47.5} & \bf\fst{60.1} & \bf\fst{1.0} \\
\bottomrule
\end{tabular}
}
\end{table*}

\begin{figure*}[h]
     \setlength{\tabcolsep}{0pt}
     \def\mywidth{.25}
     \begin{tabular}{cccc}
     \includegraphics[width=\mywidth\linewidth]{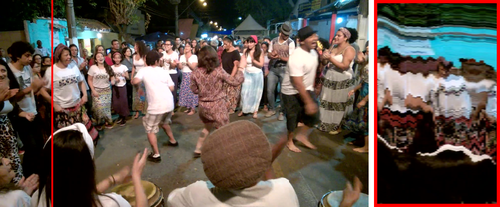} \hspace{0.1mm} &
      \includegraphics[width=\mywidth\linewidth]{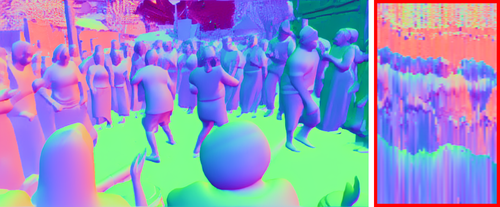} \hspace{0.1mm}&
      \includegraphics[width=\mywidth\linewidth]{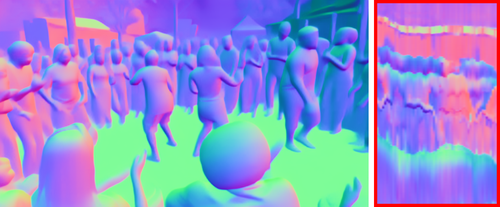} \hspace{0.1mm}&
      \includegraphics[width=\mywidth\linewidth]{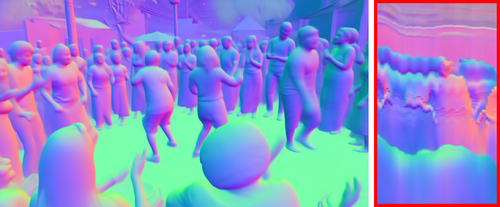}\\

      \includegraphics[width=\mywidth\linewidth]{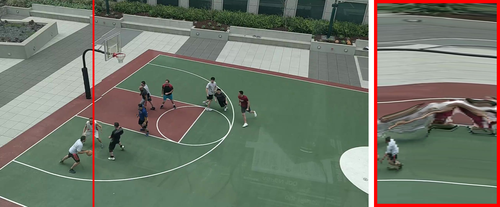} \hspace{0.1mm} &
      \includegraphics[width=\mywidth\linewidth]{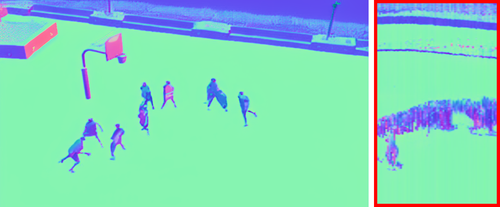} \hspace{0.1mm}&
      \includegraphics[width=\mywidth\linewidth]{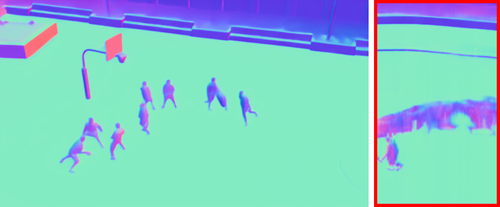} \hspace{0.1mm}&
      \includegraphics[width=\mywidth\linewidth]{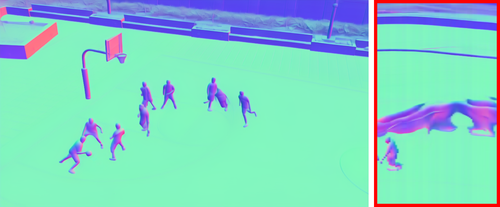}\\

      \includegraphics[width=\mywidth\linewidth]{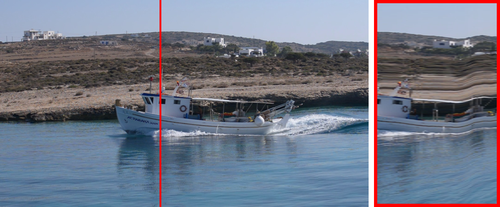} \hspace{0.1mm} &
      \includegraphics[width=\mywidth\linewidth]{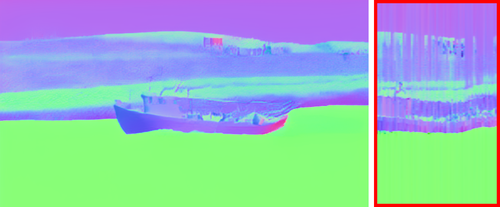} \hspace{0.1mm}&
      \includegraphics[width=\mywidth\linewidth]{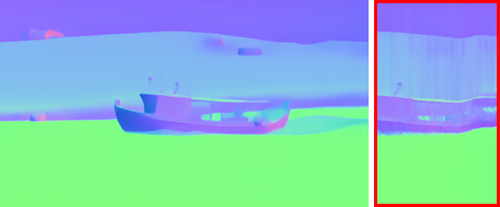} \hspace{0.1mm}&
      \includegraphics[width=\mywidth\linewidth]{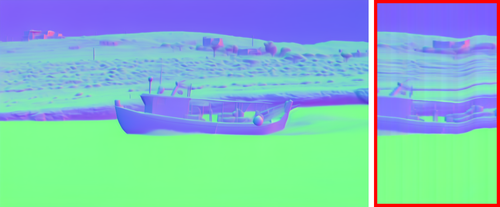}\\

      \includegraphics[width=\mywidth\linewidth]{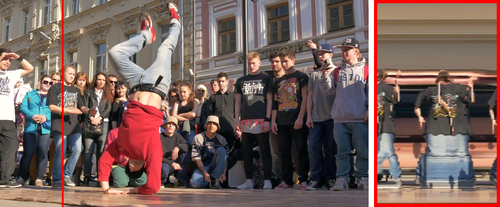} \hspace{0.1mm} &
      \includegraphics[width=\mywidth\linewidth]{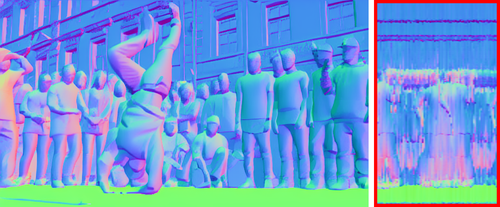} \hspace{0.1mm}&
      \includegraphics[width=\mywidth\linewidth]{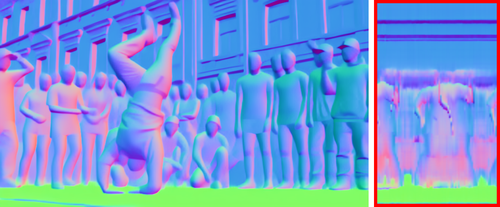} \hspace{0.1mm}&
      \includegraphics[width=\mywidth\linewidth]{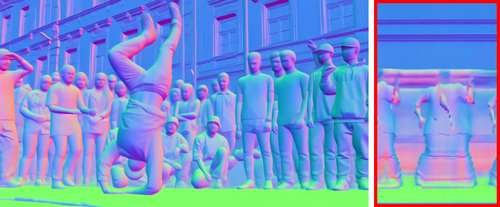}\\

      \includegraphics[width=\mywidth\linewidth]{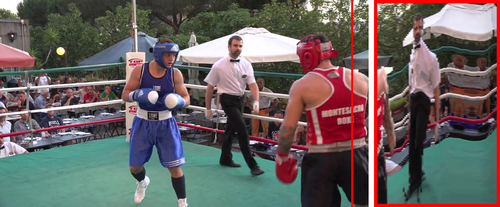} \hspace{0.1mm} &
      \includegraphics[width=\mywidth\linewidth]{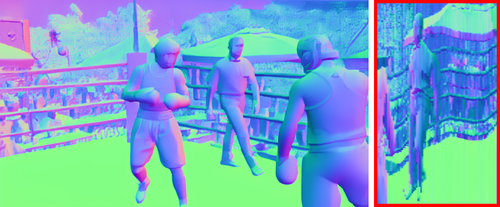} \hspace{0.1mm}&
      \includegraphics[width=\mywidth\linewidth]{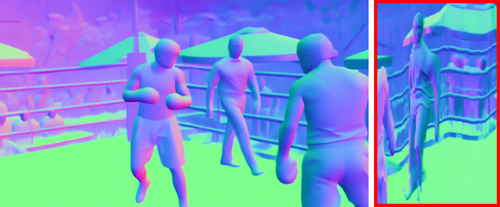} \hspace{0.1mm}&
      \includegraphics[width=\mywidth\linewidth]{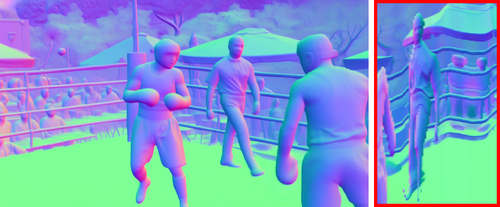}\\
      
      \includegraphics[width=\mywidth\linewidth]{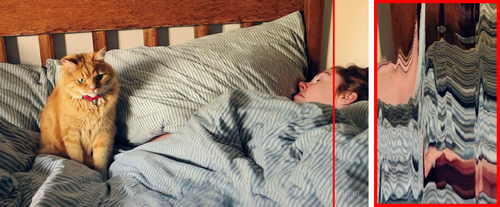} \hspace{0.1mm} &
      \includegraphics[width=\mywidth\linewidth]{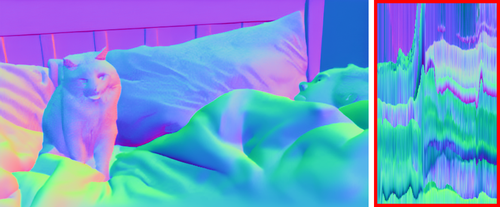} \hspace{0.1mm}&
      \includegraphics[width=\mywidth\linewidth]{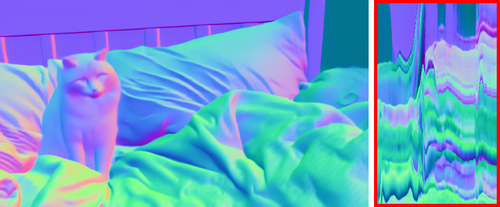} \hspace{0.1mm}&
      \includegraphics[width=\mywidth\linewidth]{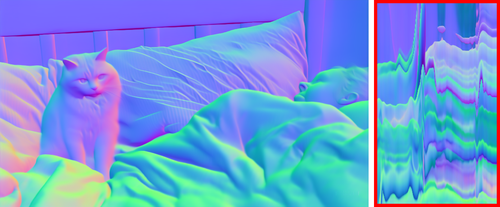}\\

      \begin{small}Input\end{small} &
      \begin{small}StableNormal\end{small} &
      \begin{small}Marigold-E2E-FT\end{small} &
      \begin{small}Ours\end{small} 
      
     \end{tabular}
     \caption{
        \textbf{Qualitative comparisons.}
        The input videos are sampled from the DAVIS dataset~\cite{davis_2019} and Sora-generated videos.
        To highlight the temporal consistency, the y-t slices at the designated red line positions are displayed in red boxes.
        }
   \label{fig:normal}
\end{figure*}

\subsection{Experimental Setup}
\noindent\textbf{Implementation details.}
We build our NormalCrafter upon the SVD~\cite{blattmann2023stable} model.
For the SFR, we resize the input images to make the DINO feature match the size of the U-Net intermediate features. $h_{\phi}$ is a three-layer perceptron while $\bm{h}_{l}$ is the output features of the second up blocks of U-Net's decoder.
We fine-tune the VAE for 20,000 iterations employing a base learning rate of $1\times 10^{-5}$.
For the U-Net, we train the first stage for 20,000 iterations using a learning rate of $3\times 10^{-5}$ and subsequently conduct end-to-end fine-tuning for 10,000 iterations with $1\times10^{-5}$ learning rate at the second stage. In the first stage, we use a 
 hybrid approach: with probability 0.5, we set the noise level $\sigma_t$ to a fixed value of 700; otherwise, we sample $\sigma_t$ from a noise level distribution $p(\sigma) = \mathcal{N}(0.7, 1.6)$ following SVD. In both stages, we resize the short edge of input clips to 576 without changing the aspect ratio.
All training processes utilize the AdamW optimizer with an exponential learning rate decay strategies following a 100-step warm-up.
We conduct training on eight GPUs with a total batch size of eight.
The U-Net training spans approximately 1.5 days, while VAE fine-tuning requires about one day.

\noindent\textbf{Training datasets.}
Following~\cite{stablenormal}, we train our model using five meticulously selected datasets, encompassing both single-frame and video types, each with high-resolution frames and ground-truth normal maps from synthetic environments.
For single-frame datasets, we utilize 49,494 images from Replica~\cite{straub2019replica} for indoor scenes and 45,620 frames from 3D Ken Burns~\cite{3d_ken_burns} for outdoor scenes.
For video datasets, we employ Hypersim~\cite{hypersim}, MatrixCity~\cite{li2023matrixcity}, and Objaverse~\cite{objaverseXL} to cover indoor scenes, outdoor scenes, and object sequences, respectively.
For Hypersim, we utilize the training subset and chain frames from each scene in sequence, yielding 613 videos. We further segment them into 1,780
short clips, each containing between 30 and 60 frames for balanced sampling during training.
For MatrixCity, we draw on the training subset of the Big City scene, restructuring frames based on camera extrinsic to produce 2,316 videos, which will then be further divided into 7601 short clips. The normal maps are generated from ground-truth depth maps using cross-product-based methods~\cite{bae2024rethinking}.
For Objaverse, we render 45,081 objects under randomly sampled continuous camera trajectories with diverse lighting to form 45,081 videos.
During training, these datasets are sampled proportionally to the number of frames in order to balance the overall training process.

\subsection{Evaluations}

\noindent\textbf{Evaluation protocols.}
We thoroughly evaluate NormalCrafter on four widely recognized benchmarks: NYUv2~\cite{nyu}, iBims-1~\cite{ibim}, ScanNet~\cite{scannet}, and Sintel~\cite{sintel}.
Among these benchmarks, NYUv2 and iBims-1 cater to single-image normal estimation, whereas ScanNet and Sintel contain video sequences.
For Sintel, we adopt the consecutive-frame split from DSINE~\cite{bae2024rethinking} to assess temporal consistency across 1064 frames from 23 scenes.
For ScanNet, we sample 20 different scenes, each providing 50 continuous frames for thorough evaluation.
We adhere to the DSINE~\cite{bae2024rethinking} evaluation protocols, computing angular deviations (measured in degrees) between the estimated normal maps and their ground-truth counterparts.
We compare mean and median angular errors, where lower values indicate superior performance, and the proportion of pixels with angular errors below certain thresholds (i.e., $11.25^{\circ}$, $22.5^{\circ}$ and $30^{\circ}$), where higher values reflect greater precision.

\noindent\textbf{Baselines.}
We comprehensively evaluate the performance of NormalCrafter against six representative baselines: DSINE~\cite{bae2024rethinking}, GeoWizard~\cite{geowizard}, GenPercept~\cite{GenPercept}, StableNormal~\cite{stablenormal}, Marigold-E2E-FT~\cite{E2E}, and Lotus-D~\cite{lotus}.
DSINE stands as the leading method among all discriminative approaches, whereas StableNormal, Marigold-E2E-FT, and Lotus-D establish the frontier among diffusion-based solutions.
All of these baselines are devised primarily for single-image normal estimation.

\noindent\textbf{Quantitative comparison.}
We first quantitatively compare our model with baseline normal estimators on both single-image benchmarks (NYUv2 and iBims) and video benchmarks (ScanNet and Sintel) in~\cref{table:benchmark}.
We can observe that our NormalCrafter achieves state-of-the-art performance on all video datasets, surpassing existing approaches by a considerable margin.
Particularly on the Sintel dataset, characterized by its substantial camera motion and fast-moving objects, NormalCrafter outperforms the second-best method across all metrics, most notably improving mean angular error ($1.6^\circ$), median angular error ($1.6^\circ$), and the proportion of pixels with angular errors below $22.5^\circ$ ($2.6$) and $30^\circ$ ($3.1$).
Moreover, on the ScanNet dataset, despite its limited camera movement and static scenes, NormalCrafter still attains the highest performance.
Compared with the second-best method, Marigold-E2E-FT~\cite{E2E}, NormalCrafter yields a $0.8^\circ$ improvement in mean angular error, alongside enhancements of $1.2$ and $1.5$ in the proportions of pixels with angular errors below $22.5^\circ$ and $30^\circ$, respectively, while delivering comparable performance on median angular error and angular errors under $11.25^\circ$.
The superior performance of NormalCrafter can be attributed to our model's ability to effectively capture temporal context and SFR to extract intrinsic semantics.

Although our model is primarily designed for video normal estimation, it can also perform single-image normal estimation by setting the frame length to one.
As shown in~\cref{table:benchmark}, NormalCrafter demonstrates either state-of-the-art or competitive performance on image-based datasets, outperforming the second-best method on the NYUv2 dataset in terms of mean angular error ($0.8^\circ$), as well as the proportions of pixels with angular errors below $22.5^\circ$ ($1.5$) and $30^\circ$ ($1.6$).
On the iBims dataset, our method remains on par with other single-image normal estimation approaches.
These results demonstrate the adaptability and robust performance of NormalCrafter, as it can effectively address both video and single-image normal estimation tasks.

\noindent\textbf{Qualitative results.}
To qualitatively evaluate the performance of NormalCrafter, we compare it with StableNormal and Marigold-E2E-FT on the DAVIS dataset~\cite{davis_2019} and Sora-generated videos~\cite{liu2024sora}, as illustrated in~\cref{fig:normal}.
StableNormal is designed for robust single-image normal estimation, while Marigold-E2E-FT represents a cutting-edge normal estimator.
To more vividly illustrate the temporal consistency of the results, we profile the y-t slices for each output within red boxes, obtained by extracting normal values along the temporal axis at designated red line positions, following~\cite{hu2024depthcrafter}.
We can observe that NormalCrafter consistently yields temporally coherent normal sequences, as evidenced by the smooth y-t slices in all examined examples, whereas both StableNormal and Marigold-E2E-FT exhibit zigzag patterns, indicating flickering artifacts in their estimations.
Moreover, NormalCrafter's predictions exhibit finer-grained details compared to those of StableNormal and Marigold-E2E-FT, thanks to the SFR, which accentuates fine-grained details. More qualitative results are provided in the supplementary material.

\subsection{Ablation study}

\begin{table*}%
    \caption{
        \textbf{Ablation study.} 
        We ablate the effectiveness of Semantic Feature Regularization (SFR), Two-Stage Training strategy (w/o Stage1 and w/o Stage2), and fine-tuning VAE decoder (VAE-FT).
        }

\footnotesize
\setlength\tabcolsep{4pt}
\resizebox{\linewidth}{!}{
\begin{tabular}{l|cc|ccc|c|cc|ccc|c}
\toprule
\multirow{2}{*}{Method} 
& \multicolumn{6}{c|}{Scannet~\cite{scannet} (Video Benchmark)}
& \multicolumn{6}{c}{Sintel~\cite{sintel} (Video Benchmark)} \\
\cline{2-13}
& mean $\downarrow$ & med $\downarrow$ & {\scriptsize $11.25^{\circ} \uparrow$} & {\scriptsize $22.5^{\circ} \uparrow$} & {\scriptsize $30^{\circ} \uparrow$} & Rank$\downarrow$
& mean $\downarrow$ & med $\downarrow$ & {\scriptsize $11.25^{\circ} \uparrow$} & {\scriptsize $22.5^{\circ} \uparrow$} & {\scriptsize $30^{\circ} \uparrow$} & Rank$\downarrow$ \\
\midrule
Ours w/o VAE-FT & 
13.4 & \bf\fst{6.8} & 67.3 & 82.8 & 87.8 & 1.8 & 30.8 & \bf\fst{23.9} & 23.4 & 47.4 & \bf\fst{60.1} & 1.8\\
Ours w/o Stage1 & 
13.4 & \bf\fst{6.8} & 67.3 & 82.8 & 87.6 & 2.0 & \bf\fst{30.7} & 24.2 & 21.5 & 46.9 & 60.0 & 2.6 \\
Ours w/o Stage2 & 
14.2 & 8.1 & 63.7 & 82.0 & 87.4 & 4.8 & 31.6 & 25.3 & 19.7 & 44.6 & 57.9 & 5.0 \\
Ours w/o SFR & 
13.7 & 7.0 & 67.1 & 82.5 & 87.4 & 4.0 & 31.1 & 24.7 & 21.4 & 45.8 & 58.9 & 4.0 \\
\hline
Ours & 
\bf\fst{13.3} & \bf\fst{6.8} & \bf\fst{67.4} & \bf\fst{82.9} & \bf\fst{87.9} & \bf\fst{1.0}
& \bf\fst{30.7} & \bf\fst{23.9} & \bf\fst{23.5} & \bf\fst{47.5} & \bf\fst{60.1} & \bf\fst{1.0}\\

\bottomrule
\end{tabular}
}

    \vspace{-0.3cm}
\label{table:ablation}
\end{table*}

\begin{figure*}[t]
     \setlength{\tabcolsep}{0pt}
     \def\mywidth{.12}
     \begin{tabular}{ccccc}%
         \raisebox{1.7\normalbaselineskip}[0.0pt][0pt]{\rotatebox{90}{Input}} \hspace{0.05mm} &
         \includegraphics[height=\mywidth\linewidth]{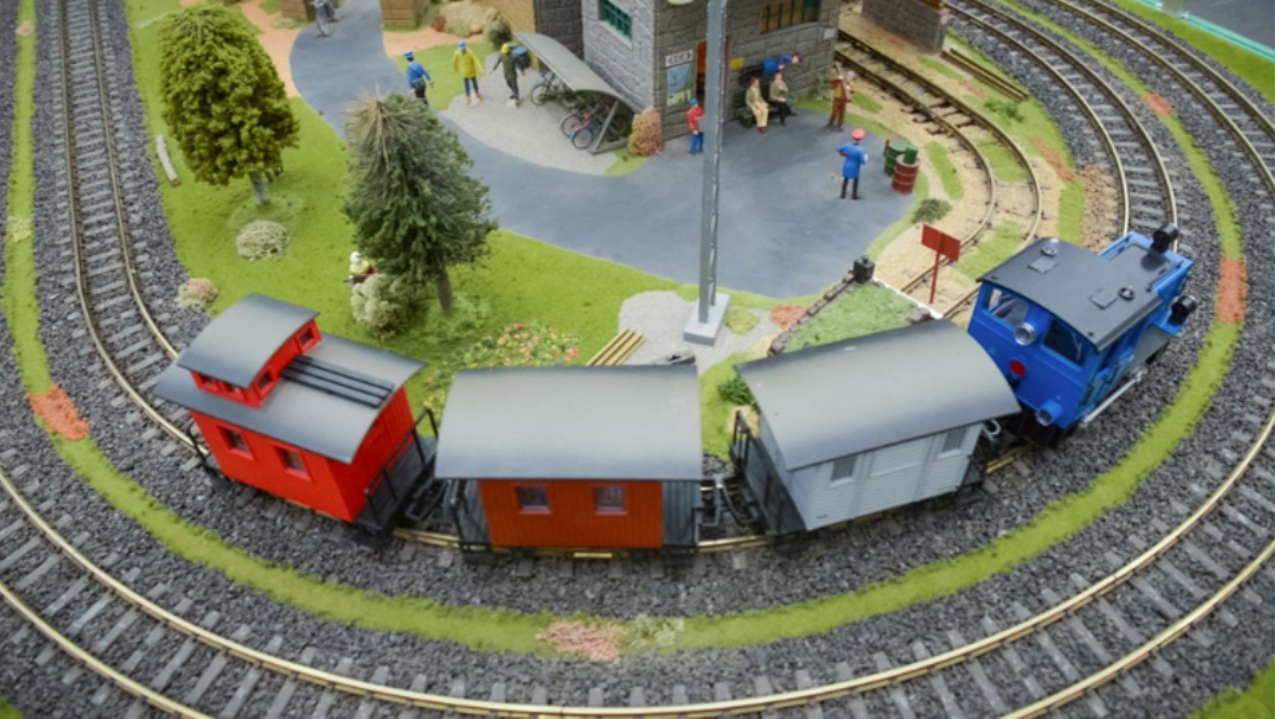} \hspace{0.05mm} &
          \includegraphics[height=\mywidth\linewidth]{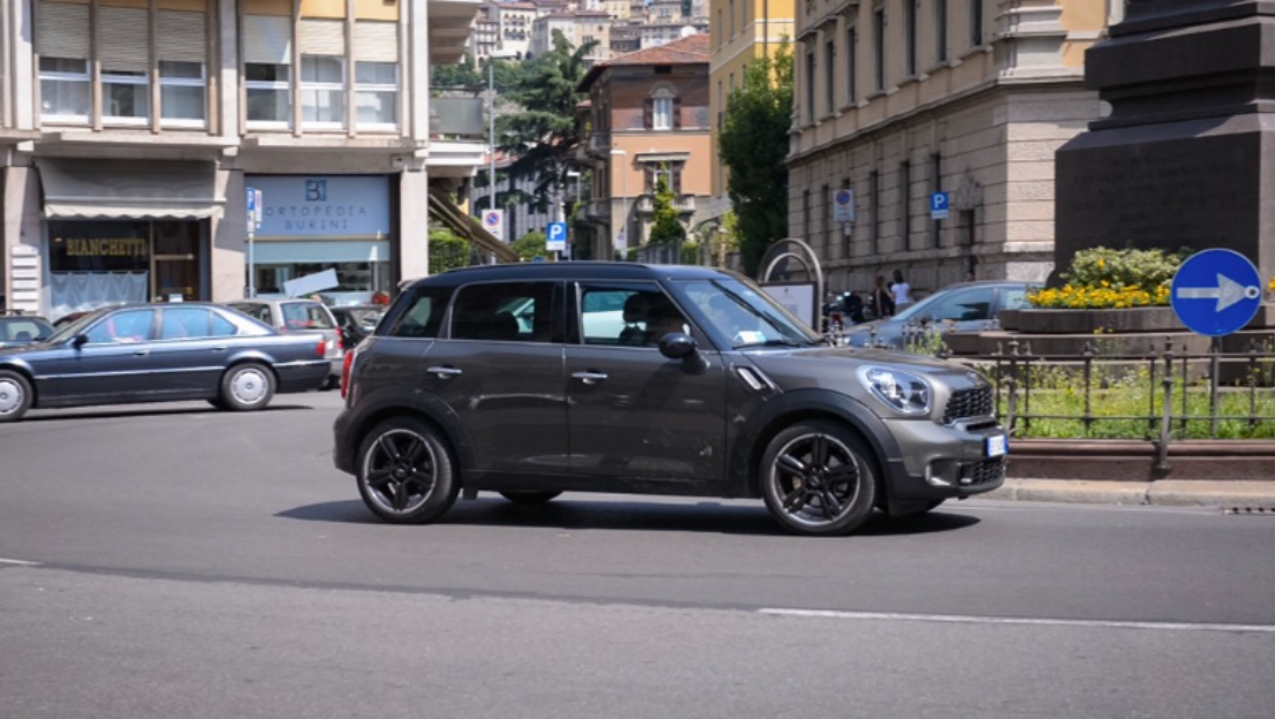} \hspace{0.05mm} &
          \includegraphics[height=\mywidth\linewidth]{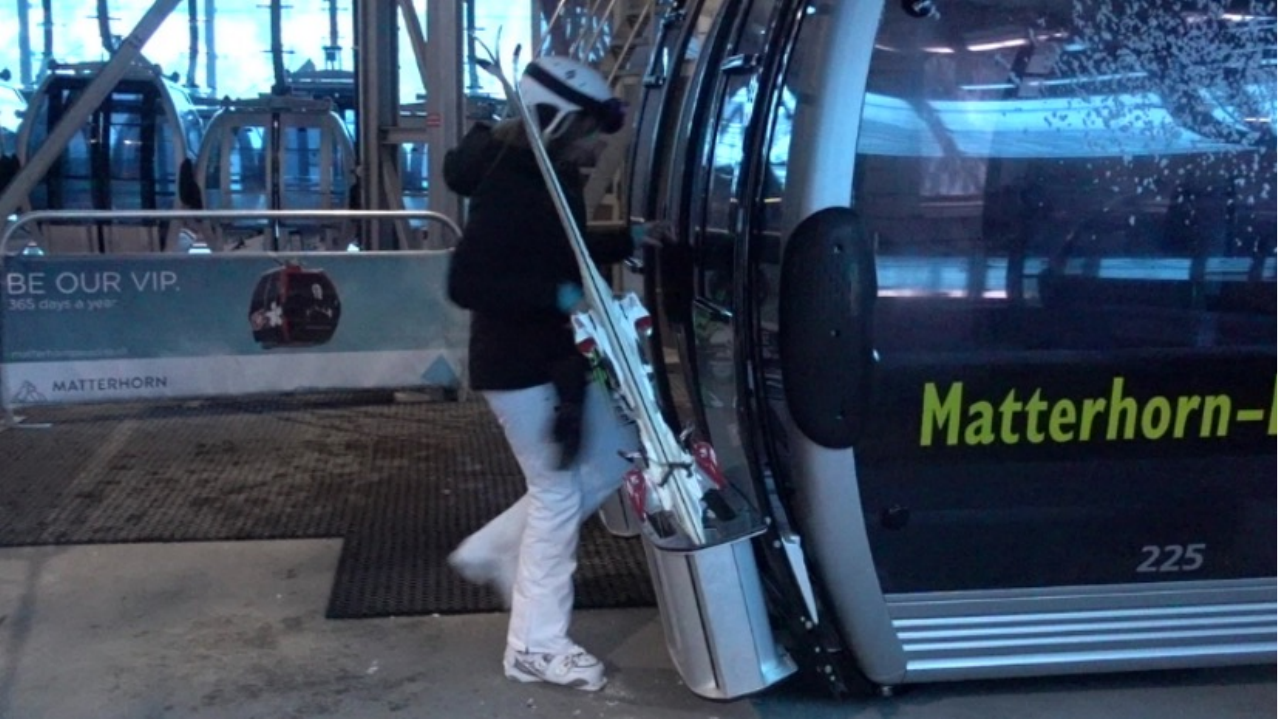} \hspace{0.05mm} &
          \includegraphics[height=\mywidth\linewidth]{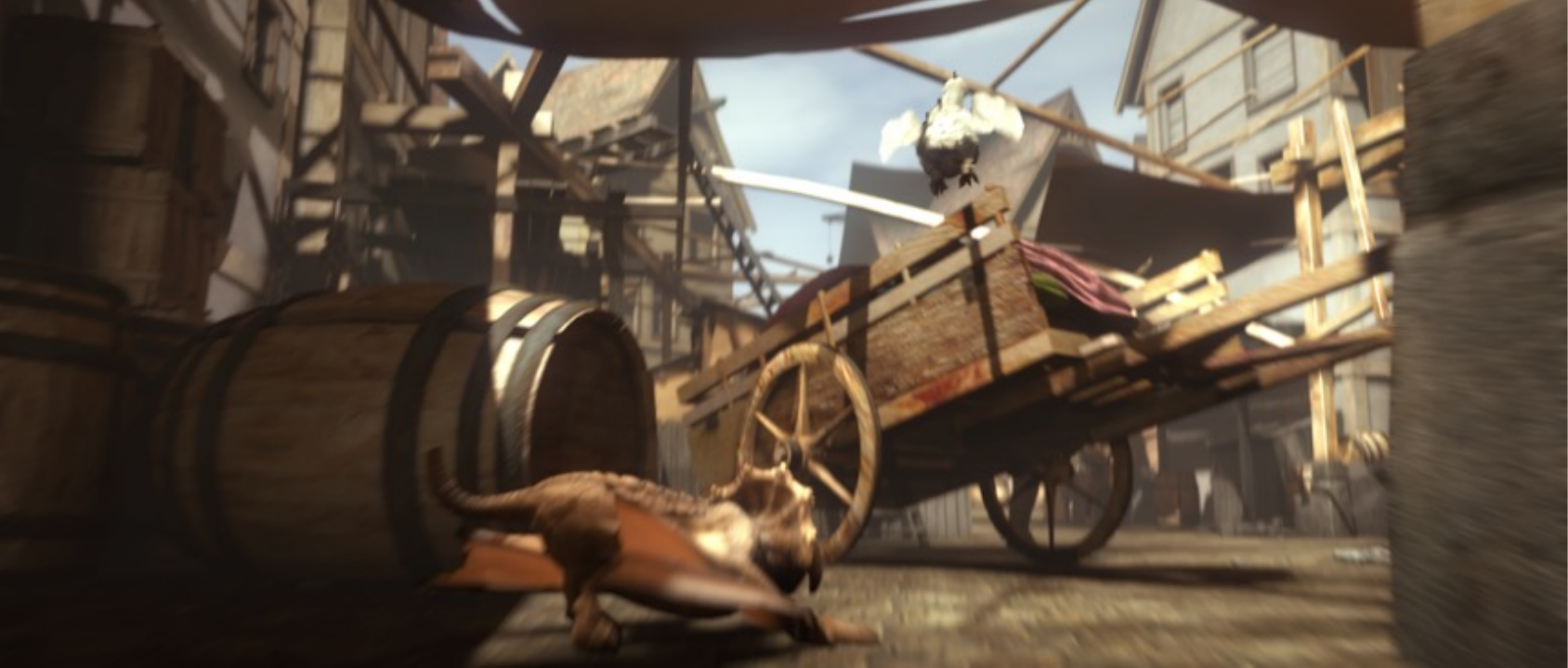} \\
          
          \raisebox{0.3\normalbaselineskip}[0.0pt][0pt]{\rotatebox{90}{Ours w/o SFR}} \hspace{0.05mm} &
          \includegraphics[height=\mywidth\linewidth]{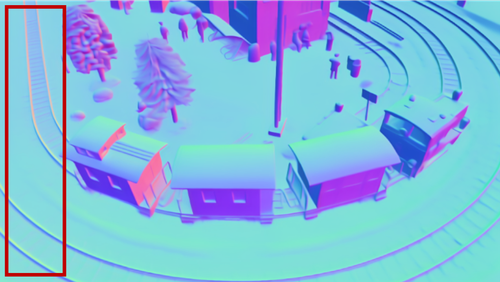} \hspace{0.05mm} &
          \includegraphics[height=\mywidth\linewidth]{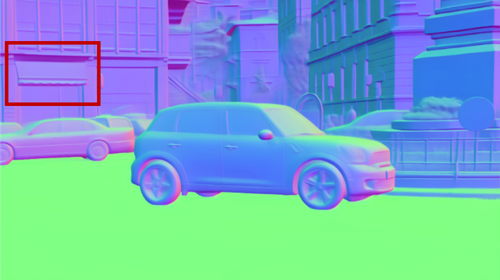} \hspace{0.05mm} &
          \includegraphics[height=\mywidth\linewidth]{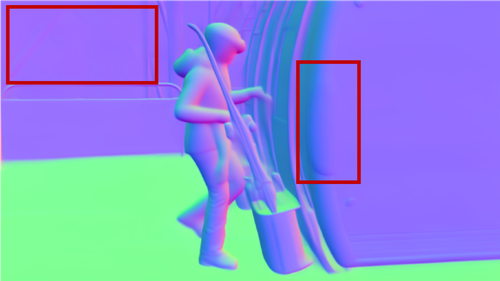} \hspace{0.05mm} &
          \includegraphics[height=\mywidth\linewidth]{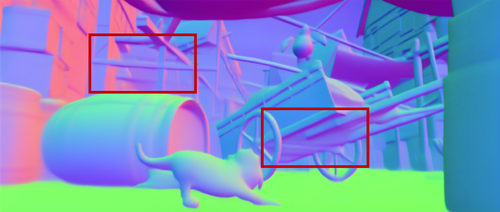} \\

          \raisebox{2.4\normalbaselineskip}[0.0pt][0pt]{\rotatebox[origin=c]{90}{Ours}} \hspace{0.05mm} &
          \includegraphics[height=\mywidth\linewidth]{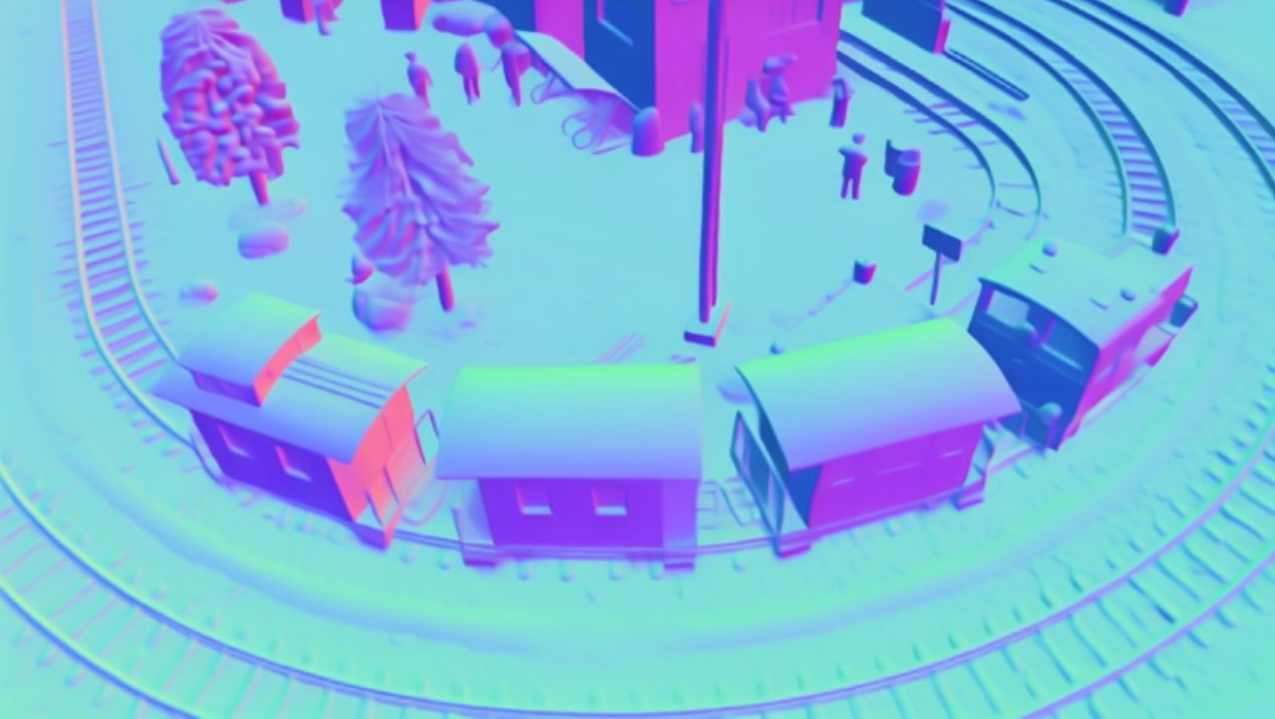} \hspace{0.05mm} &
          \includegraphics[height=\mywidth\linewidth]{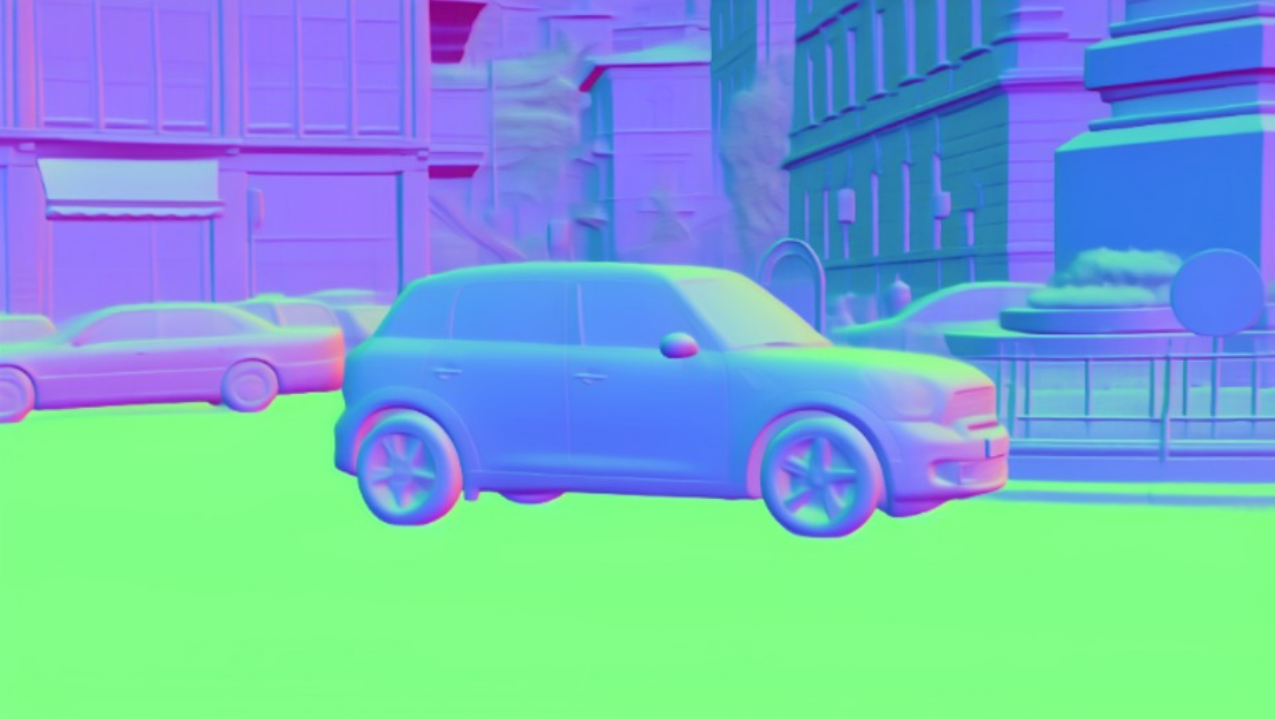} \hspace{0.05mm} &
          \includegraphics[height=\mywidth\linewidth]{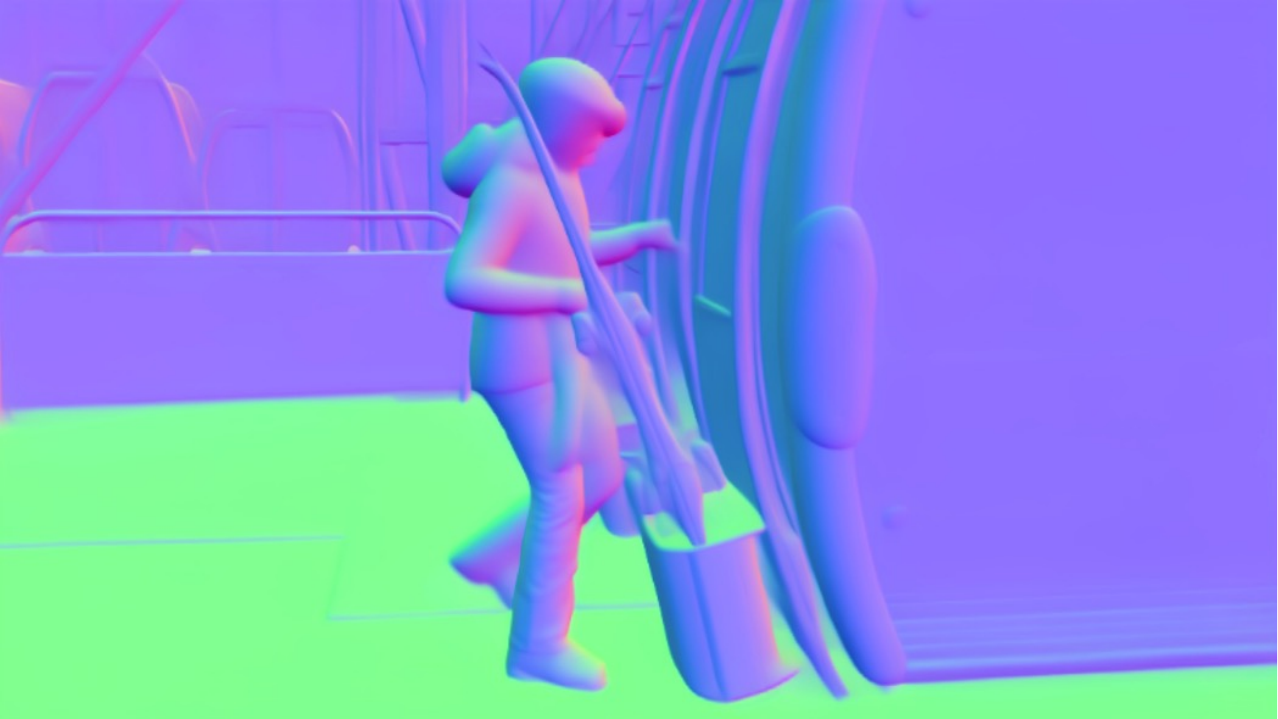} \hspace{0.05mm} &
          \includegraphics[height=\mywidth\linewidth]{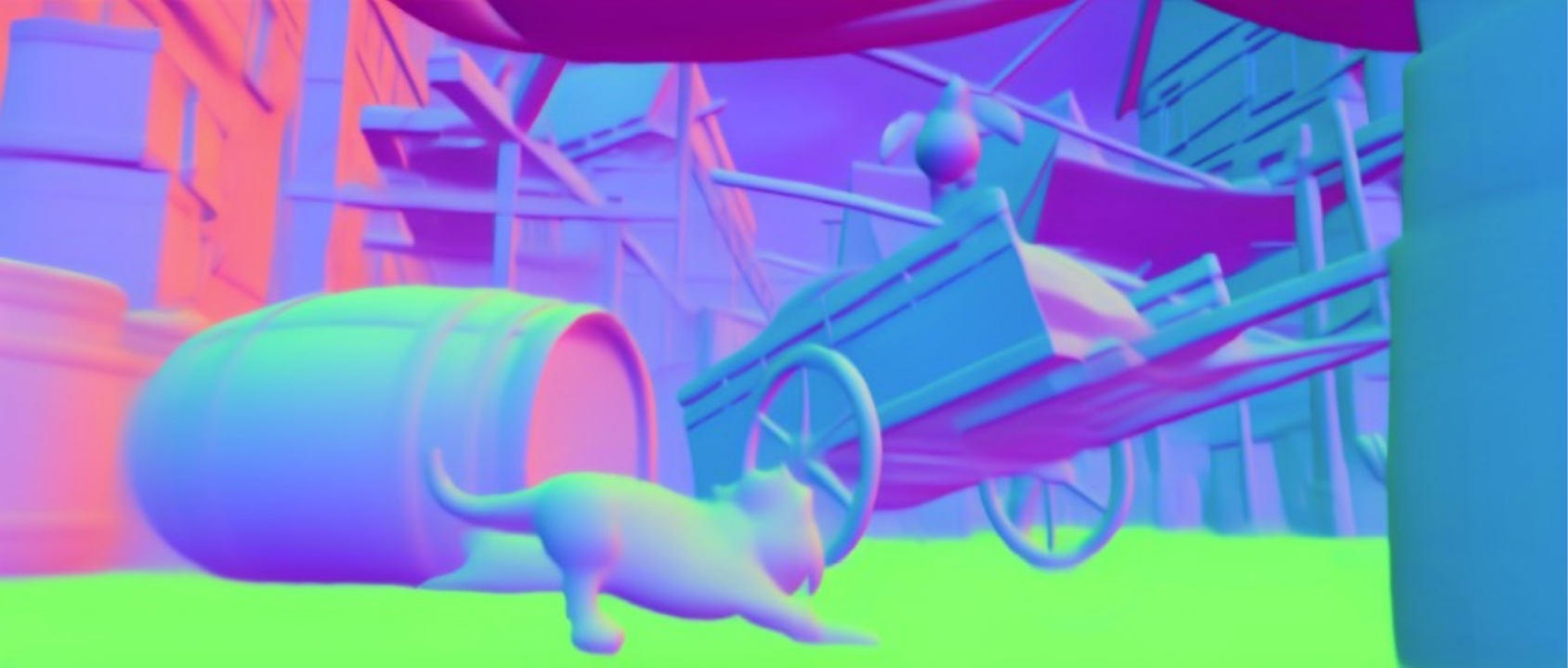} \\
     \end{tabular}
     \vspace{-0.1cm}
     \caption{
      \textbf{Ablation results} with Semantic Feature Regularization (SFR). 
      Red boxes highlight the significant differences.
      }
      \vspace{-0.1cm}
   \label{fig:ablation_SFR}
\end{figure*}

\begin{figure}[t]
     \setlength{\tabcolsep}{0pt}
     \def\mywidth{.3}
     \begin{tabular}{cccc}
         \includegraphics[height=\mywidth\linewidth]{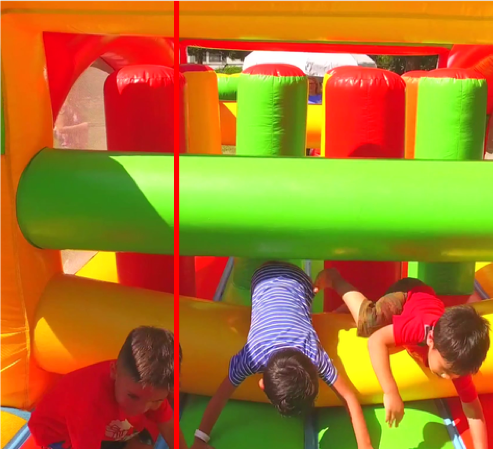} \hspace{0.05mm} &
          \includegraphics[height=\mywidth\linewidth]{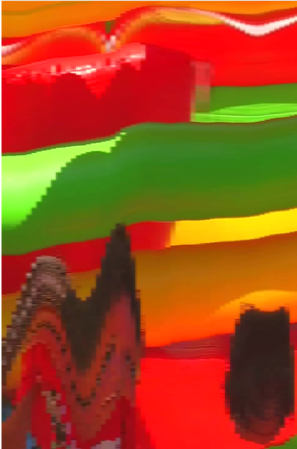} \hspace{0.05mm} &
          \includegraphics[height=\mywidth\linewidth]{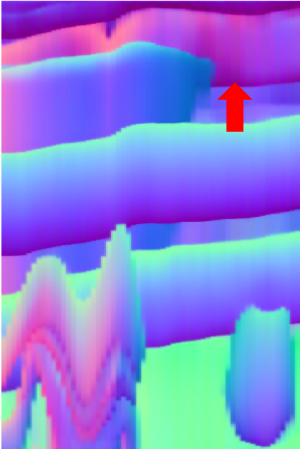} \hspace{0.05mm} &
          \includegraphics[height=\mywidth\linewidth]{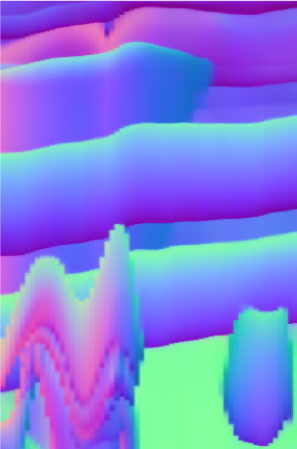} \\

          \begin{small}Input\end{small} &
          \begin{small}Input Slice\end{small} &
          \begin{small}Ours w/o Stage1\end{small} &
          \begin{small}Ours\end{small} 
     \end{tabular}
     \vspace{-0.1cm}
     \caption{
      \textbf{Qualitative Ablation Results} of two-stage fine-tuneing stategy. Without Stage1, the model suffers from temporal consistency due to the limited number of frames in training.
      }
      \vspace{-0.4cm}
   \label{fig:ablation_stage}
\end{figure}

\noindent\textbf{Effectiveness of Semantic Feature Regularization (SFR).} 
We compare the performance of NormalCrafter with and without SFR.
As shown in~\cref{table:ablation}, NormalCrafter consistently outperforms the variant without SFR across all metrics on both ScanNet and Sintel datasets.
The qualitative comparison in~\cref{fig:ablation_SFR} further illustrates the benefits of SFR, demonstrating SFR's capability to direct the diffusion model concentrate on intrinsic semantics, thereby enabling accurate and detailed normal predictions.

\noindent\textbf{Influence of SFR location.} 
The U-Net consists of four encoder blocks (``Down0-3"), one middle block (``Mid"), and four decoder blocks (``Up0-3").
We investigate the impact of SFR location by applying SFR at different layers, from ``Down1" to ``Up2".
As shown in~\cref{table:ablation_location}, the performance improvement peaks at ``Up1", indicating that the optimal location for SFR is in the middle of the network.
We suspect this is because shallow layers primarily capture low-level information, while deeper layers have too few subsequent layers to effectively map semantics to normal maps.

\noindent\textbf{Effectiveness of two-stage training.}
We ablate the effectiveness of the two-stage training strategy by training the model using stage 1 (w/o stage 2) or stage 2 only (w/o stage 1).
From~\cref{table:ablation}, the model without stage2 (w/o Stage2) performs significantly worse. On the other hand, although the model without stage1 (w/o Stage1) performs comparably with ours in spatial accuracy, it falls short in temporal consistency as shown in~\cref{fig:ablation_stage}. The above observation demonstrates that the two-stage training strategy significantly improves spatial accuracy without compromising temporal consistency.
More qualitative comparisons are provided in the supplementary material.

\noindent\textbf{Effectiveness of fine-tuning VAE.}
We evaluate the effectiveness of fine-tuning the VAE decoder. The reconstruction error of VAE decreases after fine-tuning, with mean angular error reducing from 5.75 to $\bf{4.07}$ and PSNR improving from 25.58 to $\bf{28.00}$. 
This superior decoder further positively affects the training of the normal estimator. As shown in \cref{table:ablation}, the improved performance of \textbf{Ours VAE-FT} demonstrates the effectiveness of fine-tuning VAE.

\begin{table}%
    \caption{
        \textbf{Influence of SFR location.}
        We apply SFR at different locations in the U-Net architecture, from ``Down1" to ``Up2", and analyze its impact on performance.
        }
    
    \vspace{-0.cm}

\footnotesize
\setlength\tabcolsep{3pt}
\begin{center}
\begin{tabular}{l|cc|ccc|c}
\toprule
\multirow{2}{*}{\small Method} 
& \multicolumn{6}{c}{Scannet~\cite{scannet}} \\
\cline{2-7}
& mean $\downarrow$ & med $\downarrow$ & {\scriptsize $11.25^{\circ} \uparrow$} & {\scriptsize $22.5^{\circ} \uparrow$} & {\scriptsize $30^{\circ} \uparrow$}  & Rank$\downarrow$ \\
\midrule
w/o SFR & 
13.7 & 7.0 & 67.1 & 82.5 & 87.4 & 7.0\\
Down1 & 
13.5 & 6.8 & 67.6 & 82.5 & 87.4 & 3.8\\
Down2 & 
13.6 & 6.8 & 67.2 & 82.4 & 87.3 & 6.4\\
Down3 & 
13.5 & 6.8 & 67.4 & 82.5 & 87.5 & 4.2\\
Mid & 
13.5 & 6.8 & 67.5 & 82.7 & 87.7 & 3.4\\
Up0 & 
13.4 & 6.8 & 67.5 & 82.8 & 87.8 & 2.4\\
Up1(Ours) & 
\bf\fst{13.3} & 6.8 & 67.4 & \bf\fst{82.9} & \bf\fst{87.9} & 2.0\\
Up2 & 
13.4 & \bf\fst{6.7} & \bf\fst{67.8} & \bf\fst{82.9} & 87.8 & \bf\fst{1.4}\\

\midrule
\multirow{2}{*}{\small Method} 
& \multicolumn{6}{c}{Sintel~\cite{sintel}} \\
\cline{2-7}
& mean $\downarrow$ &  med $\downarrow$ &{\scriptsize $11.25^{\circ} \uparrow$} &{\scriptsize $22.5^{\circ} \uparrow$} & {\scriptsize $30^{\circ} \uparrow$} & Rank$\downarrow$  \\
\midrule
w/o SFR & 
31.1 & 24.7 & 21.4 & 45.8 & 58.9 & 6.2\\
Down1 & 
31.0 & 24.3 & 21.4 & 46.6 & 59.6 & 3.8\\
Down2 & 
31.2 & 24.9 & 21.2 & 45.7 & 58.7 & 7.8\\
Down3 & 
31.1 & 24.5 & 21.2 & 46.2 & 59.3 & 6.0 \\
Mid & 
31.0 & 24.3 & 21.5 & 46.6 & 59.7 & 3.2\\
Up0 & 
\bf\fst{30.7} & \bf\fst{23.8} & 21.5 & \bf\fst{47.5} & \bf\fst{60.5} & 1.4\\
Up1(Ours) & 
\bf\fst{30.7} & 23.9 & \bf\fst{23.5} & \bf\fst{47.5} & 60.1 & \bf\fst{1.4}\\
Up2 & 
31.1 & 24.3 & 21.7 & 46.7 & 59.5 & 3.6\\
\bottomrule
\end{tabular}
\end{center}

    \vspace{-0.3cm}

\label{table:ablation_location}
\end{table}

\subsection{Limitations}
Although our method achieves the state-of-the-art performance in terms of spatial accuracy and temperal consistency in video normal estimation, its large parameter size poses challenges for deployment on mobile devices. Therefore, optimizing the model’s efficiency through model pruning, model quantization and distillation techniques could be a potential direction for future work.
\section{Conclusion}
We present NormalCrafter, a video normal estimator that can generate temporally consistent normal sequences with fine-grained details for open-world videos. The temporal consistency is achieved by leveraging video diffusion priors, while the spatial accuracy with details is enhanced by semantic feature regularization. Additionally, a two-stage training strategy further improved spatial accuracy while maintaining long temporal context by leveraging both
latent and pixel space learning. Extensive evaluations have demonstrated that NormalCrafter achieves state-of-the-art performance in open-world video normal estimation under zero-shot settings. We hope our work can provide inspiration for future investigations in this domain.

% \clearpage
{
    \small
    \bibliographystyle{ieeenat_fullname}
    \bibliography{main}
}

\end{document}